\documentclass[letterpaper, 10 pt, journal, twoside]{IEEEtran}

\usepackage{graphicx} 
\usepackage{epsfig} 
\usepackage{mathptmx} 
\usepackage{times} 
\usepackage{amsmath} 
\usepackage{amssymb}  
\usepackage{bm}
\usepackage{hyperref}
\usepackage[capitalize]{cleveref}
\usepackage{array}
\usepackage{multirow}
\usepackage{tikz}
\usepackage{url}
\usepackage[T1]{fontenc}
\usepackage{comment}
\usepackage{todonotes}
\usepackage{enumitem}

\title{Aleatoric Uncertainty from AI-based 6D Object Pose Predictors for Object-relative State Estimation}
\author{Thomas Jantos$^{1}$, Stephan Weiss$^{1}$ and Jan Steinbrener$^{1}$

\thanks{Manuscript received: February, 27, 2025; Revised June, 25, 2025; Accepted August, 29, 2025.}
\thanks{This paper was recommended for publication by Editor Markus Vincze upon evaluation of the Associate Editor and Reviewers' comments.}
\thanks{This work was supported by the Austrian Ministry of Climate Action and Energy (BMK), grant agreement FO999895366 (EMFLanding).}
\thanks{$^{1}$The authors are with the Control of Networked Systems Group, University of Klagenfurt, 9020 Klagenfurt am Wörthersee, Austria {\tt\footnotesize \{name.surname\}@ieee.org}}%
\thanks{{\textbf{Pre-print version, accepted Aug/2025, DOI follows ASAP ~\copyright IEEE.
}}}
}

\begin{document}
\maketitle
\begin{abstract}
Deep Learning (DL) has become essential in various robotics applications due to excelling at processing raw sensory data to extract task specific information from semantic objects. For example, vision-based object-relative navigation relies on a DL-based 6D object pose predictor to provide the relative pose between the object and the robot as measurements to the robot's state estimator. Accurately knowing the uncertainty inherent in such Deep Neural Network (DNN) based measurements is essential for probabilistic state estimators subsequently guiding the robot's tasks. Thus, in this letter, we show that we can extend any existing DL-based object-relative pose predictor for aleatoric uncertainty inference simply by including two multi-layer perceptrons detached from the translational and rotational part of the DL predictor. This allows for efficient training while freezing the existing pre-trained predictor. We then use the inferred 6D pose and its uncertainty as a measurement and corresponding noise covariance matrix in an extended Kalman filter (EKF). Our approach induces minimal computational overhead such that the state estimator can be deployed on edge devices while benefiting from the dynamically inferred measurement uncertainty. This increases the performance of the object-relative state estimation task compared to a fix-covariance approach. We conduct evaluations on synthetic data and real-world data to underline the benefits of aleatoric uncertainty inference for the object-relative state estimation task.
\end{abstract}



\vspace{-0.5cm}
\section{Introduction}

Deep neural networks (DNNs) excel at computer vision tasks such as object detection, classification, and 6D object pose prediction. The latter is significant for various robotics applications as it provides metric information about the robot's relative position and orientation with respect to the objects in the scene. Possible robotics applications include robotic manipulation and object-relative navigation with uncrewed aerial vehicles (UAV) for, e.g., infrastructure inspection. Even recent work \cite{jantos2024aivio,jantos2023ai}, however, does not include the corresponding uncertainty in this process. Indeed, a DNN that can capture its aleatoric uncertainty to account for and quantify the inherent noise in the data would enable more robust predictions, better decision-making under uncertainty, and appropriate weighting of unreliable inputs in downstream tasks.
\looseness=-1

An extended Kalman filter (EKF) statistically combines information from multiple sources, e.g., sensors, to estimate the system's state. The measurement covariance quantifies the uncertainty in sensor observations and determines how much weight the filter gives to the measurements compared to the predicted state during the update step. A higher measurement covariance reduces the influence of the measurements, while lower covariance increases their impact, ensuring that the filter adapts appropriately to the reliability of the data.

While EKF covariance tuning for optimized estimator performance is often done and possible for experts, specific single use cases, and for specific measurement sensors, it is hardly an approach scalable and resilient to non-experts under different situations. Also, such tuning becomes more sophisticated when DNNs instead of physical sensors are used to generate the measurements. Recently, Jantos et al. \cite{jantos2024aivio,jantos2023ai} showed with such tuning that 6D object poses predicted by a DNN from raw RGB images can be fused with inertial measurement unit (IMU) data for localization of a mobile robot with respect to objects of interest. Given these advances in DNNs for object-relative navigation, quantifying the uncertainty inherent in the DNN's prediction is more important than ever to better determine the measurement's significance and increase the estimation performance. 
\looseness=-1

\begin{figure}
    \centering
    \includegraphics[width=1.0\linewidth]{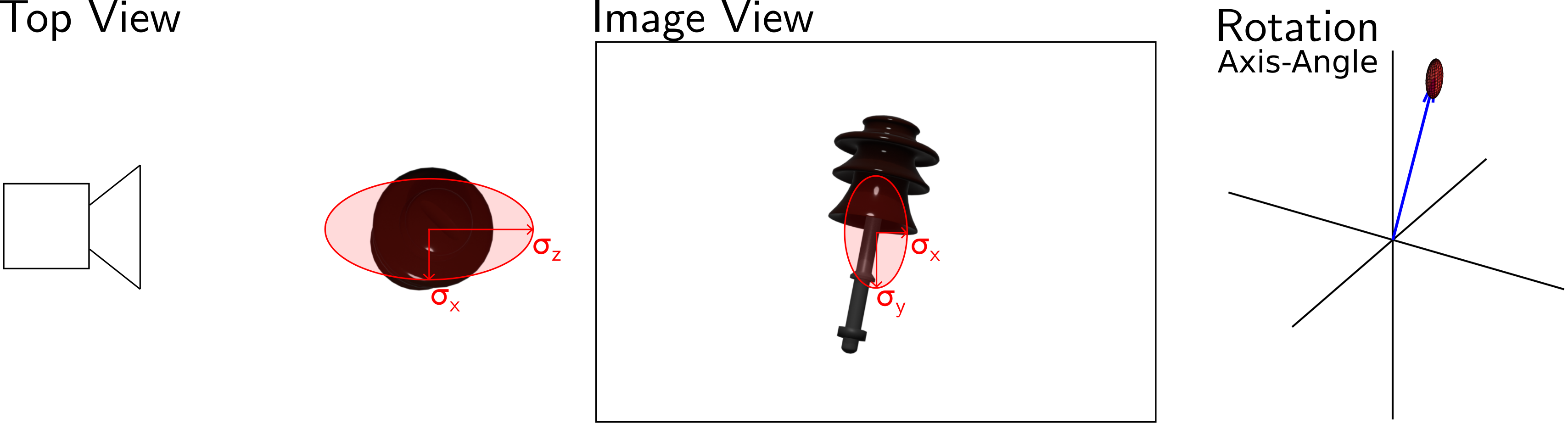}
    \caption{Exemplary visualization of the predicted aleatoric uncertainty of the 6D pose. The network predicts the uncertainty separately for position and for the rotation. In the axis-angle representation this uncertainty can be interpreted as the uncertainty about the axis of rotation, i.e. vector direction, and the angle of rotation, i.e. length of vector. We visualize the 1-$\sigma$ bound of the 3D Gaussian uncertainty. Please note, that the visualized uncertainties are overemphasized for better understanding.}
    \label{fig:example_aleatoric_uncertainty}
    \vspace{-0.7cm}
\end{figure}

In this letter, we demonstrate that aleatoric uncertainty can be utilized as measurement covariance in EKFs, combining the best of both worlds, i.e., the high accuracy of the DNN with the statistical signal weighting of an EKF, and replacing the need for any expert knowledge or manual tuning of the measurement uncertainty parameters. We show that a (pre-trained) 6D object pose predictor can be extended to additionally infer the metric, aleatoric uncertainty of the predicted poses, see \cref{fig:example_aleatoric_uncertainty}. Replacing the fixed measurement covariance found with experts' tuning efforts in the EKF update step with the dynamically predicted aleatoric uncertainty yields advantages from an ease-of-use point of view and leads to performance improvements. It also allows for confidence based reference object switching and outlier rejection in multi-object relative navigation tasks.
\looseness=-1

Even though the work presented here will be discussed in the context of inspection with resource-constrained UAVs, the insights gained can be translated into other (object-relative) state estimation scenarios. Our contributions are the following:
\begin{itemize}[leftmargin=*,noitemsep,topsep=0pt,parsep=0pt,partopsep=0pt]
    \item Extending a (pre-trained) 6D object pose predictor for aleatoric uncertainty inference. Modeling the aleatoric uncertainty as multivariate Gaussians to capture the uncertainty of the full 6D pose.
    \item Utilizing the predicted metric uncertainty as a dynamic measurement covariance matrix in an EKF for object-relative state estimation.
    \item Implementing uncertainty-based dynamic reference object switching and aleatoric uncertainty-based outlier rejection (AOR) for improved multi-object-relative state estimation.
    \looseness=-1
    \item Validating the proposed contributions on synthetic and real-world data.
\end{itemize}

\noindent The remainder of the letter is structured as follows. In \cref{sec:related_work}, we summarize the related work regarding uncertainty prediction in deep learning and its application in state estimation. In \cref{sec:method}, we present our 6D object pose predictor capable of aleatoric uncertainty prediction and the integration of aleatoric uncertainty as dynamic measurement covariance in our state estimator for multi-object-relative state estimation. In \cref{sec:experiments}, the experiments and corresponding results are discussed. In \cref{sec:concluison}, the letter is concluded.
\vspace{-0.2cm}
\section{Related Work} \label{sec:related_work}

With increased prevalence of deep learning approaches across various fields, the importance of quantifying and managing uncertainties of DNN's predictions has grown significantly \cite{gawlikowski2023aisurvey}. The are mainly two types of uncertainty: aleatoric uncertainty, i.e., noise inherent in the data, and epistemic uncertainty, describing the lack of knowledge of the model. While aleatoric uncertainty can not be reduced, epistemic uncertainty can be lowered by increasing the training data size/diversity. For aleatoric uncertainty, an underlying error distribution is assumed and the network is trained to predict these parameters \cite{kendall2017uncertainties}. 
\looseness=-1

In object pose prediction, uncertainty prediction refers to the prediction of a pose distribution. Bingham distributions can be used to model the orientation distribution \cite{sato2023probabilistic}. Alternatively, non-parametric distributions are used to implicitly model the pose distribution \cite{haugaard2023spyropose} or ensembles are used to capture the pose uncertainty \cite{wursthorn2024uncertainty}. Merrill et al. \cite{merrill2022symmetry} predict the 2D-pixel uncertainty in keypoint-based object pose prediction and utilize these uncertainties as a covariance for graph-based, object-relative simultaneous localization and mapping (SLAM). Instead of keypoint uncertainty, Zorina et al. \cite{zorina2025temporally} empirically determine a fixed 6D pose uncertainty over a test dataset to use in object-relative SLAM. However, the 6D pose covariance is simplified to use the same variance for $x$ and $y$ and a single variance value for the three rotational components. 

Reliable and accurate state estimation, essential for deploying mobile robots in complex environments and performing various tasks, relies on combining information from multiple sources, e.g., different sensor measurements. In D3VO, Yang et al. \cite{yang2020d3vo} trained a neural network for depth, pose, and pixel brightness uncertainty predictions and used them in the optimization process of a visual-inertial odometry (VIO) algorithm. NVINS \cite{han2024nvins} is a VIO framework that utilizes a DL-based camera pose regressor and IMU measurements in factor-graph optimization. Additionally, the network predicts aleatoric and epistemic uncertainty of the pose, modeled as a three-dimensional Gaussian for the translation and one-dimensional, isotropic Langevin distribution for the rotation. Combining these two uncertainties results in a four-dimensional covariance matrix, which functions as a weighting factor for the pose measurements during the optimization. Peretroukhin et al. \cite{peretroukhin2019probabilistic} train a multi-head network for aleatoric and epistemic rotation uncertainty prediction, modeled as a three-dimensional Gaussian in the Lie algebra tangent to the predicted rotation. The predicted rotation and its corresponding uncertainty are used in graph-based optimization for visual odometry (VO). Russell and Reale \cite{russell2021multivariate} discuss the possibility of using DL-based, multivariate aleatoric and epistemic uncertainty as measurement noise in an EKF for two use cases. First, 3D object tracking by predicting the 3D position and uncertainty. Second, VO by predicting the change in angle and position and the corresponding two-dimensional uncertainty. CoordiNet \cite{moreau2022coordinet} is a DNN that predicts the camera pose and aleatoric uncertainty from an input image. While the translation uncertainty is assumed to be a three-dimensional Gaussian, the network predicts a single uncertainty value for the rotation, from which the three-dimensional covariance matrix is constructed.
\looseness=-1

In contrast to the above work, our approach models the full 6D object pose uncertainty as a multivariate Gaussian with individual parameters for the components. Thus in this letter, we propose a method to extend any 6D object pose prediction network to predict aleatoric uncertainty modeled as multivariate Gaussians. To the best of our knowledge, we are the first to utilize the predicted aleatoric uncertainty of the full 6D pose as a dynamic measurement noise covariance matrix in an EKF for object-relative state estimation.
\vspace{-0.2cm}

\section{Method} \label{sec:method}
In this section, we first describe the deep-learning pose prediction framework, its extension for aleatoric uncertainty prediction, and the loss function. Second, we present the state estimation framework and the integration of the predicted aleatoric uncertainty as a dynamic measurement covariance matrix.
\looseness=-1

\subsection{Aleatoric Uncertainty for 6D Object Pose Prediction}
\label{subsec:object_pose_estimation}

A wide variety of deep learning-based approaches exist for 6D object pose prediction, differing in the input modality, availability of additional information, and possible post-processing steps. In this work, we show that a DNN that directly predicts the 6D pose of an object can be easily extended for aleatoric uncertainty prediction. As an example, we choose the open source available object pose prediction framework PoET \cite{jantos2023poet}. It does not rely on any additional input information and directly predicts the relative 6D pose of each object from a single RGB image. Epistemic uncertainty will not be considered as we train the object pose predictor with synthetic data, meaning enough data can be generated to cover the object's $SO(3)$ space. Besides that, the additional computational load introduced, caused by multiple forward passes, would not be suitable for mobile robot navigation, potentially impeding its real-time implementation as the same authors showed in \cite{jantos2024aivio}.
\looseness=-1

\begin{figure}
    \centering
    \includegraphics[width=1.0\linewidth]{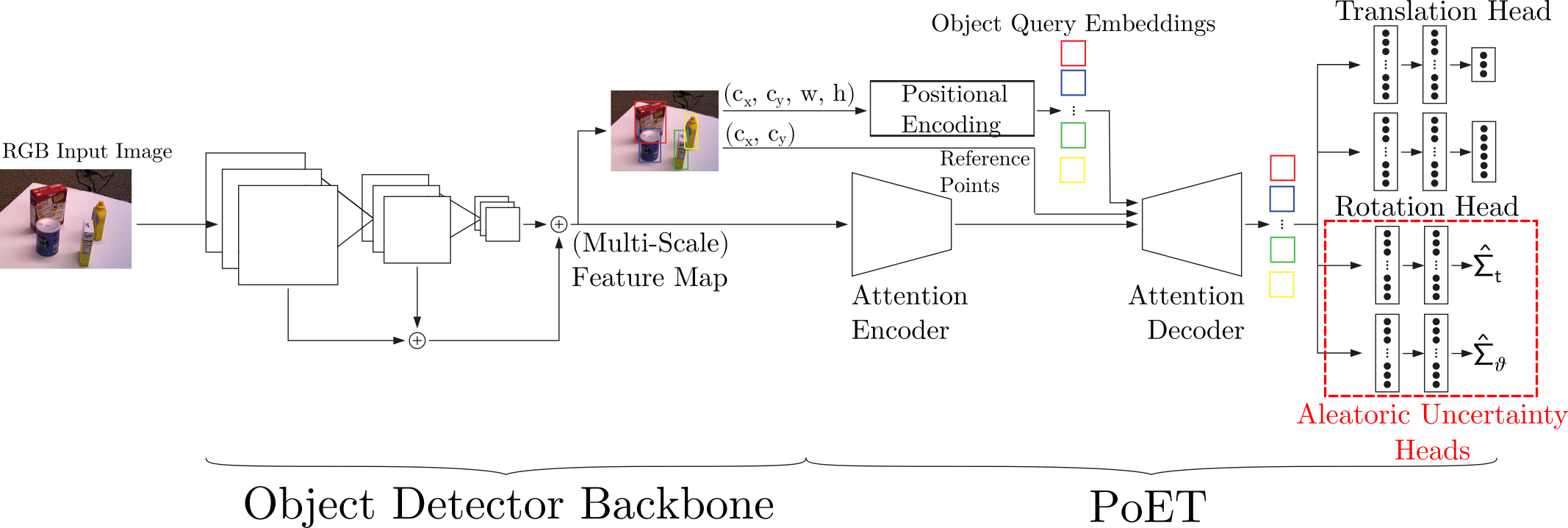}
    \caption{Our proposed extension to the network architecture of PoET \cite{jantos2023poet} for aleatoric uncertainty prediction (\texttt{red box}). 
    Using the object queries, the aleatoric heads predict the corresponding uncertainties $\hat{\Sigma}_{t,\vartheta} \in \mathbb{R}^{3x3}$.}
    \label{fig:aleatoric_poet}
    \vspace{-0.8cm}
\end{figure}

After the RGB image is passed through an object detection backbone, the detected objects and multi-scale feature maps are fed into a deformable transformer to predict object queries containing object-specific and global image context information. These object queries are passed separately to a translation and rotation head, simple multi-layer perceptrons (MLPs), to predict each object's 3D translation and 6D rotation \cite{zhou2019continuity}. Using Gram-Schmidt, the 6D rotation is transformed into $\textbf{R} \in SO(3)$ of which we take the matrix logarithm to receive an element of its Lie algebra, which is a skew-symmetric matrix
\vspace{-0.5cm}
\begin{equation}
    log(\mathbf{R}) = \theta [v]_\times \in \mathfrak{so}(3),
\end{equation}
where $\theta$ is the angle and $v$ the axis of rotation. From this we can get the axis-angle representation of the rotation with $\vartheta = \theta v \in \mathbb{R}^3$. We extend the network by two additional MLPs to predict each object's aleatoric uncertainty for the translation $\hat{\Sigma}_t$ and rotation $\hat{\Sigma}_\vartheta$. The aleatoric uncertainty heads predict a class-specific uncertainty in a multi-class scenario. The network architecture is visualized in \cref{fig:aleatoric_poet}.

The main advantages of adding dedicated aleatoric uncertainty heads, i.e., being detached from the translation and rotation head, is \emph{the possibility to extend any pre-trained object pose prediction framework for uncertainty inference} and that the heads are able to encapsulate uncertainty-specific features in the MLP's weights. While the network is still end-to-end trainable, \emph{the uncertainty heads can be trained individually by freezing the rest of the network,} reducing the training time drastically. Moreover, the additional computational overhead is minimal during deployment, thus making our approach suitable for edge device deployment and mobile robotics.

Following a similar idea as \cite{peretroukhin2019probabilistic} and \cite{russell2021multivariate}, we model the aleatoric uncertainty as being data dependent and a multivariate Gaussian distribution with the mean given by the model output. Given an input image $x$, our network outputs the translation $\hat{t}$ and its covariance $\hat{\Sigma}_t$, $\mathcal{N}(\hat{t}, \hat{\Sigma}_t)$, as well as the rotation $\hat{\vartheta}$ and its covariance $\hat{\Sigma}_\vartheta$, $\mathcal{N}(\hat{\vartheta}, \hat{\Sigma}_\vartheta)$, with
\begin{equation}
\label{eq:aleatoric_cov}
    \hat{\Sigma}_t = \begin{pmatrix}
\hat{\sigma}_x^2 & 0 & 0\\
0 & \hat{\sigma}_y^2 & 0 \\
0 & 0 & \hat{\sigma}_z^2
\end{pmatrix}
 \text{ and }
 \hat{\Sigma}_\vartheta = \begin{pmatrix}
    \hat{\sigma}_{\vartheta_1}^2 & 0 & 0\\
    0 & \hat{\sigma}_{\vartheta_2}^2 & 0 \\
    0 & 0 & \hat{\sigma}_{\vartheta_3}^2
\end{pmatrix}.
\end{equation}
We predict the uncertainty for the individual components of the predicted translation and rotation and assume no cross-correlation between them. As for the predicted object poses, the uncertainties are expressed with respect to the camera frame (see \cref{fig:example_aleatoric_uncertainty} for an explanatory example). While the translational uncertainty refers to the variance of the predicted position, the rotational uncertainty is expressed in the Lie algebra tangential to the predicted rotation.

Considering $\hat{t}$ and $\hat{\vartheta}$ to be independent of each other, we assume that the probability of the output $y \in \{t, \vartheta\}$ can be modeled by a multivariate Gaussian distribution. Given that the network output $\hat{y}$ for the input image $x$ models the mean, $\mathbb{E}[y|x]$, the likelihood is then
\begin{equation}
\label{eq:probability}
    p(y|x) = (2\pi)^{-3/2} |\hat{\Sigma}_y|^{-1/2} \text{exp}\left(-\frac{1}{2} (y-\hat{y})^T \hat{\Sigma}_y^{-1} (y-\hat{y}) \right).
\end{equation}
Taking the negative log-likelihood of \cref{eq:probability}, we can define the loss function for a single object $n$ to be
\begin{align}
    \mathcal{L}_{y_n} &= \frac{1}{2} (y_n-\hat{y}_n)^T \hat{\Sigma}_{y_n}^{-1} (y_n-\hat{y}_n) + \frac{1}{2} \text{ln}|\hat{\Sigma}_{y_n}|\\
    &= \frac{1}{2} e_n^T \hat{\Sigma}_{y_n}^{-1} e_n + \frac{1}{2} \text{ln}|\hat{\Sigma}_{y_n}|,
\end{align}
where $e_n=(e_{n,1}, e_{n,2}, e_{n,3})^T$ defines the error between components of the ground truth $y_n$ and predicted value $\hat{y}_n$ of object $n$. 
In combination with our independent aleatoric heads, this loss definition allows us to either simultaneously or separately train the network for predicting $\hat{y}_n$ and the corresponding covariance $\hat{\Sigma}_{y_n}$. Given our assumption that $\hat{\Sigma}_{y_n}= \text{diag}(\hat{\sigma}_{y_{n,1}}^2, \hat{\sigma}_{y_{n,2}}^2, \hat{\sigma}_{y_{n,3}}^2)\in\mathbb{R}^{3x3}$, the loss function simplifies to
\begin{equation}
\label{eq:loss_function}
    \mathcal{L}_{y_n} = \frac{1}{2} \left( \sum_{i=1}^3 \frac{e_{n,i}^2}{\hat{\sigma}_{y_{n,i}}^2} + \text{ln}(\hat{\sigma}_{y_{n,i}}^2)\right).
\end{equation}
The sum of the scaled, squared error components is related to the squared Euclidean norm of a vector. For the translation, the loss function indirectly minimizes the Euclidean distance between the ground truth and predicted translation vectors. The Euclidean norm of the axis-angle is equal to the angle $\theta$ of the rotation. Hence, the loss function indirectly minimizes the geodesic distance between the ground truth and predicted rotation. Inspired by \cite{kendall2017uncertainties}, we predict $s_{y_{n,i}}=\text{ln}(\hat{\sigma}_{y_{n,i}}^2)$ for better numerical stability during training, resulting in
\begin{equation}
    \mathcal{L}_{y_n} = \frac{1}{2} \left( \sum_{i=1}^3 \text{exp}(-s_{y_{n,i}}) e_{n,i}^2 + s_{y_{n,i}}\right).
\end{equation}

To address multiple objects in an image and arbitrary batch sizes, the loss is averaged by the number of objects $N$. Finally, the network is trained with a weighted multi-task loss.
\begin{equation}
    \mathcal{L}_y = \frac{1}{N} \left(\sum_{n=1}^N\mathcal{L}_{y_n}\right) \text{ and } \mathcal{L} = \lambda_t \mathcal{L}_t + \lambda_\vartheta \mathcal{L}_\vartheta \quad .
\end{equation}
\vspace{-1.0cm}
\subsection{Object-relative State Estimation}
\label{subsec:state_estimation}

In object-relative state estimation, the current state of a mobile robot, in particular the position and orientation of its propagation sensor, the IMU ($I$), is estimated with respect to objects of interest ($O_k$) in the world ($W$) (i.e., the navigation frame) by measuring the relative 6D pose of the objects. Given two coordinate frames $A$ and $B$, the transformation of frame $B$ with respect to frame $A$ is defined by the translation ${\mathbf{p}}_{\scriptscriptstyle AB}$ and rotation ${\mathbf{R}}_{\scriptscriptstyle AB}$. $\mathbf{I}$ and $\mathbf{0}$ refer to the identity and the null matrix, respectively. Alternatively, a rotation is expressed by a quaternion  ${\mathbf{q}}_{\scriptscriptstyle AB} = [\mathbf{q_v}~q_w]^T = [q_x ~ q_y ~ q_z ~ q_w]^T$. While optimization-based methods usually achieve more accurate state estimates, they are computationally more demanding as they optimize over multiple sensor measurements. In contrast, filter-based methods, such as the EKF, are better suited for mobile robotics as they are computationally more efficient. Hence, we choose MaRS \cite{brommer2020mars} as our state estimation framework as it was developed with mobile robotics and modularity in mind. This allows for the straightforward implementation of a object-relative pose sensor with a dynamic measurement covariance.
\looseness=-1

Building on top of previous work \cite{jantos2024aivio, jantos2023ai}, the relative pose of objects predicted from RGB images by a DL-based object pose predictor is used for object-relative state estimation. Under the assumption that $N$ objects of interest are present in the scene, the state vector $\mathbf{X}$ is given by
\vspace{-0.4cm}

\footnotesize
\begin{equation}
 \mathbf{X} = [{\mathbf{p}}_{\scriptscriptstyle WI}^T, {\mathbf{v}}_{\scriptscriptstyle WI}^T, {\mathbf{q}}_{\scriptscriptstyle WI}^T, {\mathbf{b}}_{\scriptscriptstyle \omega}^T, {\mathbf{b}}_{\scriptscriptstyle a}^T, {\mathbf{p}}_{\scriptscriptstyle IC}^T, {\mathbf{q}}_{\scriptscriptstyle IC}^T,{\mathbf{p}}_{\scriptscriptstyle O_0W}^T, {\mathbf{q}}_{\scriptscriptstyle O_0W}^T, \dots, {\mathbf{p}}_{\scriptscriptstyle O_NW}^T, {\mathbf{q}}_{\scriptscriptstyle O_NW}^T]^T.
\end{equation}
\normalsize
The core states for state propagation are the position ${\mathbf{p}}_{\scriptscriptstyle WI}$, velocity ${\mathbf{v}}_{\scriptscriptstyle WI}$ and orientation ${\mathbf{q}}_{\scriptscriptstyle WI}$ of the IMU, the gyroscopic bias ${\mathbf{b}}_{\scriptscriptstyle \omega}$ and the accelerometer bias ${\mathbf{b}}_{\scriptscriptstyle a}$. We estimate the calibration between the IMU and the camera given by ${\mathbf{p}}_{\scriptscriptstyle IC}$ and ${\mathbf{q}}_{\scriptscriptstyle IC}$. Additionally, we estimate the objects in the world (${\mathbf{p}}_{\scriptscriptstyle O_kW}, {\mathbf{q}}_{\scriptscriptstyle O_kW}$) to relate the object frames to the navigation frame.

In general, an EKF assumes the measurement to depend on a non-linear measurement function of the state corrupted by an additive zero-mean Gaussian noise with the measurement noise covariance matrix $\bm{\Sigma}$, representing the uncertainty of the measurement. Modeling this measurement noise covariance matrix accurately is essential to the estimation process as it is part of the innovation covariance matrix $\textbf{S}$, which captures the uncertainty of the innovation and thus quantifies the trustworthiness of the measurement in the state update step. We will refer to the measurement noise covariance matrix as measurement uncertainty for better readability throughout the manuscript. Determining $\bm{\Sigma}$ requires either information about sensor specifications from the manufacturer, an expert to conduct empirical estimation, or iteratively fine-tuning the noise parameters. Empirical analysis of the error distribution on a validation set is a non-learning-based method to determine the measurement uncertainty of a DNN.
\looseness=-1

Taking the partial derivative of \cref{eq:loss_function} with respect to one of the uncertainty components $\hat{\sigma}_{y_{n,i}}$ and given that the standard deviation is strictly positive ($\hat{\sigma}_{y_{n,i}} > 0$), it can be shown that the minimum of the loss function is reached for $\hat{\sigma}_{y_{n,i}}^2 = e_{n,i}^2$. Therefore, during training, the aleatoric head's parameters will be optimized so that the predicted uncertainty captures the error of the network. By directly training our object pose estimation network to output a metric uncertainty of its prediction, we can eliminate the need for expert knowledge and manual fine-tuning of the measurement uncertainty in the EKF. Instead, as the aleatoric uncertainty is predicted per image and object, it varies over time and thus allows for a dynamic measurement uncertainty.

For each image, PoET detects all objects of interest and predicts their relative 6D pose to the camera (${\mathbf{p}}_{\scriptscriptstyle CO_k}, {\mathbf{q}}_{\scriptscriptstyle CO_k}$). Referring to \cite{jantos2023ai}, the measurement needs to be inverted to be able to derive the observation matrix $\textbf{H}_{O_k}$. We treat the predicted aleatoric uncertainties, see \cref{eq:aleatoric_cov}, as the measurement uncertainty for the translation $\bm{\Sigma}_t$ and rotation $\bm{\Sigma}_\vartheta$ measurements, respectively.

$\bm{\Sigma}_t$ is a $3\times3$ matrix and quantifies the measurement noise for the $xyz$-components. According to \cite{sola2017quaternion}, in quaternion-based error-state Kalman filters, small rotational disturbances $\Delta\vartheta\in\real\mathbb{R}^3$ can be specified in the local vector space tangent to the actual rotation $\mathbf{R}_{\scriptscriptstyle CO_k}\in SO(3)$. Hence, the measurement uncertainty $\bm{\Sigma}_\vartheta$ of these disturbances can be expressed in the Lie algebra by a regular $3\times3$ matrix. Exactly as for the pose measurements, the predicted uncertainties are expressed in the camera frame and hence need to be inverted using the error propagation law.
\looseness=-1
\begin{equation}
 \bm{\Sigma}_{y,\scriptscriptstyle O_kC} = \mathbf{R}_{\scriptscriptstyle O_kC} \bm{\Sigma}_{y,\scriptscriptstyle CO_k} \mathbf{R}_{\scriptscriptstyle O_kC}^T \quad .
\end{equation}
The full measurement noise covariance matrix $\bm{\Sigma}_{\scriptscriptstyle O_k}\in\mathbb{R}^{6x6}$ for object $O_k$ is constructed by diagonally stacking the inverted translation and rotation measurement noise matrices:
\begin{equation}
\label{eq:full_measurement_cov}
    \bm{\Sigma}_{\scriptscriptstyle O_k} = \begin{pmatrix}
        \bm{\Sigma}_{t, \scriptscriptstyle O_kC} & \textbf{0}_{3\times3} \\
         \textbf{0}_{3\times3} & \bm{\Sigma}_{\vartheta, \scriptscriptstyle O_kC}      
    \end{pmatrix} \quad .
\end{equation}
To render object-relative state estimation observable, it is necessary to introduce an anchor object ($O_A$), whose pose in the world frame is fixed (thus anchoring the navigation frame to its pose) \cite{jantos2023ai}. Dynamic measurement uncertainty, enabled through the predicted aleatoric uncertainty, allows for a stochastic informed decision to seamlessly switch the anchor object during the estimation process. By summing up the individual components of the predicted aleatoric uncertainty, \cref{eq:aleatoric_cov}, the object with the lowest uncertainty, i.e., the measurement the network is most certain about, is chosen as the anchor object for the current update step. We will show below, that dynamically switching the anchor object based on predicted measurement uncertainty will improve the performance of the object-relative state estimator. In case of a fixed measurement uncertainty, dynamic switching of the anchor object is not possible as the filter has no notion about the currently most certain measurement. Note that each anchor object switching process introduces a slight global drift according to the current global estimation error of the new anchor object when switched to. From an object-relative navigation perspective, this is tolerable since object relative motion is still ensured despite the global drift. In addition, switching to the currently most certain object as anchor heavily reduces outliers to be erroneously included in the estimation.

\vspace{-0.3cm}
\section{Experiments \& Results} \label{sec:experiments}
In this section, we present the experimental setup and results. We evaluate the performance of the 6D object pose predictor on a test dataset, analyze its aleatoric uncertainties and demonstrate how these aleatoric uncertainties improve object-relative state estimation.
\begin{figure}
    \centering
    \includegraphics[width=0.45\linewidth]{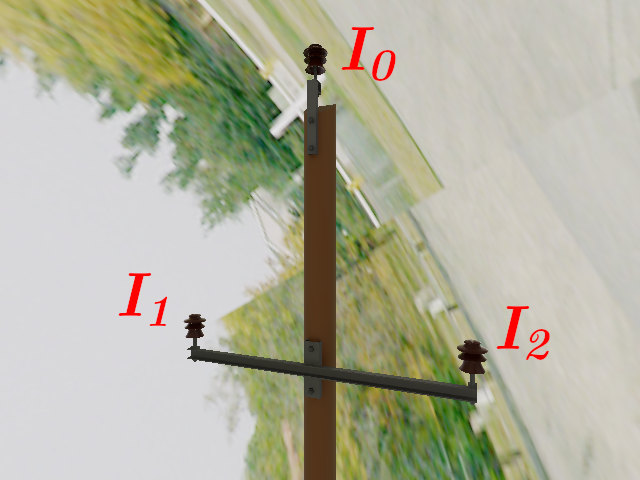}
    \includegraphics[width=0.45\linewidth]{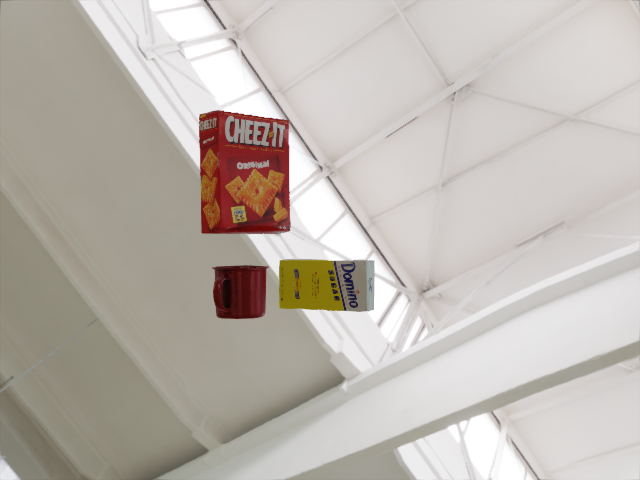}
    \vspace{-0.2cm}
    \caption{Example of synthetically generated images for the power pole (\texttt{left}) and the YCB-V objects (\texttt{right}). The numeration of the insulators used during the discussion of the results is indicated in red.}
    \label{fig:example_synt_image}
    \vspace{-0.5cm}
\end{figure}
\vspace{-0.3cm}
\subsection{Dataset \& Network Details}
\label{subsec:dataset}
We utilize NVIDIA Omniverse IsaacSim to generate synthetic images of power poles with the insulators being automatically annotated, as they serve as our objects of interest in the object-relative state estimator. Similarly, we create a synthetic dataset with a subset of the YCB-V objects \cite{xiang2018posecnn}. In both cases, our training and validation dataset consists of 100,000 and 20,000 images, respectively. Synthetic example images are shown in \cref{fig:example_synt_image}. We choose Scaled-YOLOv4 \cite{wang2021scaledyolo} as our object detection backbone and PoET's transformer consists of five encoder and decoder layers with 16 attention heads. The translation and rotation heads, as well as both aleatoric uncertainty heads are simple MLPs with an input layer, one hidden layer, and an output layer. Network training is performed on a NVIDIA GeForce RTX 3090.

In order to evaluate the state estimator's performance, we also simulate physically feasible UAV trajectories with synthetic IMU data (200 Hz) and corresponding synthetic images (15 Hz). The generated trajectories include varying heights and distances to the power pole and YCB-V objects.

The additional computational load of the aleatoric heads is measured for a NVIDIA Jetson Orin AGX 64GB DevKit. Optimizing with TensorRT 8.5.2 with full computational precision, the average image processing time for PoET without aleatoric heads amounts to 106.73ms. The processing time increases to 107.43ms for PoET with aleatoric heads, thus a mere increase of 0.6\%.
\looseness=-1

\vspace{-0.3cm}
\subsection{Aleatoric Uncertainty for 6D Object Pose Prediction}
\label{subsec:uncertainty_eval}

In \cref{subsec:object_pose_estimation}, it was highlighted that the proposed aleatoric uncertainty heads could be either trained separately or in an end-to-end fashion. To investigate the difference, we train two different networks. First, we train PoET without the aleatoric uncertainty heads as described in \cite{jantos2023poet} for 50 epochs. To evaluate the performance of the trained network, we calculate the Euclidean and geodesic distance between the ground truth and predicted translation and rotation. On the validation dataset, the trained network achieves an average translation and rotation error of 3cm and $2^\circ$, respectively. To calibrate the network for aleatoric uncertainty estimation, we freeze the network, add the aleatoric heads, and train with the loss function described in \cref{subsec:object_pose_estimation} for 10 epochs (\texttt{Calib}). As the rest of the network is frozen, the training time for an epoch decreases drastically, and the average errors do not change. Second, we train the network for simultaneous 6D object pose and aleatoric uncertainty prediction for 50 epochs (\texttt{E2E}). It achieves the same performance as the \texttt{Calib} network on the synthetic validation data.
\begin{figure}
    \centering
    \includegraphics[width=0.8\linewidth]{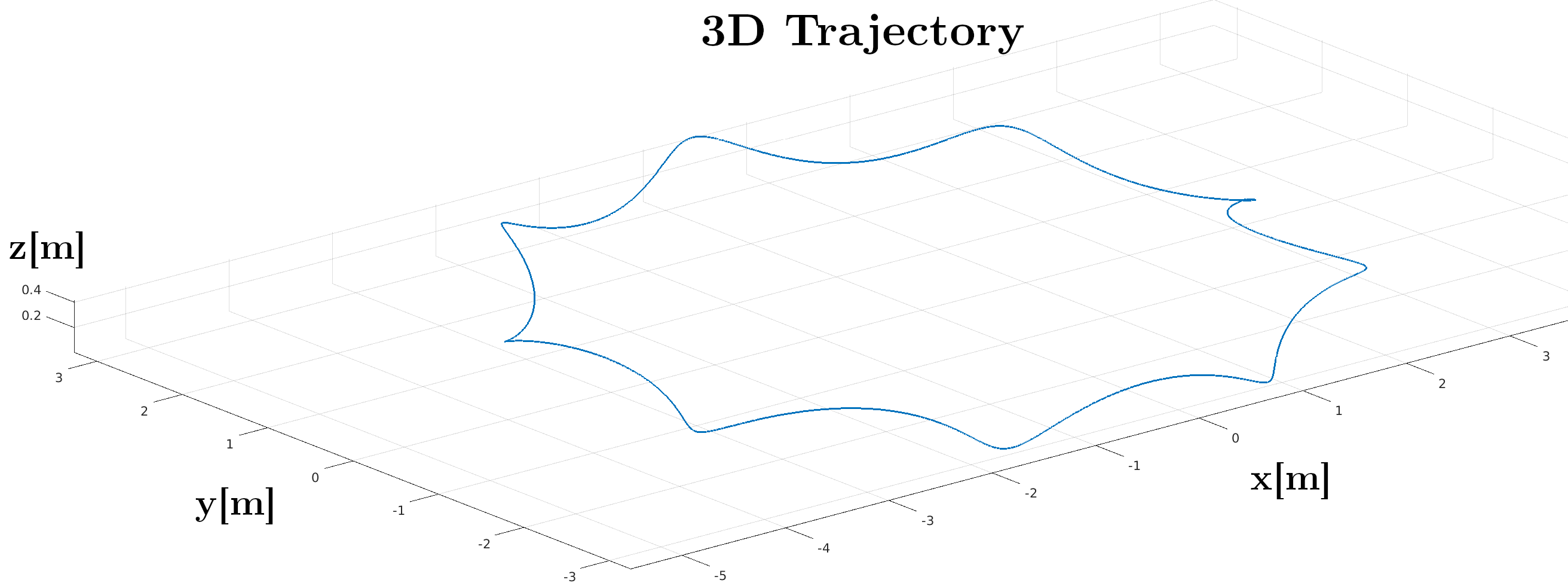}
    \caption{Visualization of the synthetic trajectory 1 used for analysis in \cref{subsec:uncertainty_eval}. While rotating to keep at the center located power pole in the view, the UAV also varies its height and distance to the power pole.}
    \label{fig:example_synt_trajectory}
    \vspace{-0.5cm}
\end{figure}

To analyze the predicted uncertainty by these two models, we take a synthetically generated trajectory as described in \cref{subsec:dataset}. On this trajectory, see \cref{fig:example_synt_trajectory}, the distance to the power pole varies between 2.7m and 3.4m, while the height follows a sinusoidal with an amplitude of 0.2m. Due to the varying distance to the power pole, the change of height of the simulated UAV and the trajectory covering $360^\circ$, the appearance of the power pole in the images covers a wide variety of viewpoints, ideal for the uncertainty analysis. Given the ground truth information from the simulation environment, it is possible to calculate the error statistics and perform the uncertainty analysis across the whole trajectory for each insulator. In addition to the translation and rotation error described above, we also calculate the absolute error component-wise for the translation and the axis-angle representation of the rotation. Similarly, the predicted aleatoric uncertainty is evaluated for each individual component. Given our assumption that the predicted 6D pose and aleatoric uncertainty describe the mean and variance of a Gaussian distribution, the quality of the uncertainty is given by the prediction interval coverage probability (PICP) \cite{gawlikowski2023aisurvey} defined as
\vspace{-0.4cm}

\footnotesize
\begin{equation}
    \text{PICP} = \frac{1}{D} \sum_{d=1}^{D} \mathbb{I} \left( y_d \in \left[\hat{y}_d - \frac{1}{2} \Phi^{-1}(\alpha) \cdot \hat{\sigma}_d, \; \hat{y}_d + \frac{1}{2} \Phi^{-1}(\alpha) \cdot \hat{\sigma}_d \right] \right),
\end{equation}

\normalsize
where $D$ is the number of data points, $\mathbb{I}$ is the indicator function, $y_d$ is the true value, $\hat{y}_d$ is the predicted value, $\hat{\sigma}_d$ is the predicted standard deviation, $ \Phi^{-1}$ is the inverse of the cumulative distribution function of the standard normal distribution and $\alpha$ is the confidence level. It describes the proportion of true values that lie within the predicted confidence intervals.


In \cref{tab:traj_10_picp}, we report the PICP metric. Assuming a confidence level of 95\%, both networks achieve a PICP of above 0.9 for each component, which is a positive indicator for well-calibrated uncertainty predictions. However, PICP values close to 1 indicate that the network overestimates the uncertainty, i.e., making the predictions more uncertain than they are. As will be shown in \cref{subsec:eval_state_estimation}, the predicted uncertainties are reliable enough to improve object-relative state estimation compared to fixed covariance values. To show that it is sufficient to calibrate a pre-trained 6D object pose predictor for aleatoric uncertainty prediction, the subsequent uncertainty analysis and the experiments for object-relative state estimation will be conducted using the \texttt{Calib} network.


\begin{table}[t]
    \centering \caption{Comparison of the PICP score per insulator for $\alpha=0.95$}
\begin{tabular}{c|c|c|c||c|c|c}
 & \multicolumn{3}{c||}{\texttt{Calib}} & \multicolumn{3}{c}{\texttt{E2E}}\tabularnewline

 & $I_0$ & $I_1$ & $I_2$ & $I_0$ & $I_1$ & $I_2$\tabularnewline
\hline 
\hline 
x & 1.00 & 0.99 & 0.99 & 1.00 & 0.99 & 1.00\tabularnewline

y & 0.99 & 1.00 & 1.00 & 0.99 & 0.99 & 0.96\tabularnewline

z & 0.97 & 0.95 & 0.84 & 0.93 & 0.97 & 0.93\tabularnewline
\hline 
\hline 
$\vartheta_1$ & 0.96 & 0.95 & 0.98 & 0.93 & 0.94 & 0.95\tabularnewline

$\vartheta_2$ & 0.99 & 0.99 & 0.96 & 1.00 & 1.00 & 1.00\tabularnewline

$\vartheta_3$ & 1.00 & 1.00 & 0.99 & 0.93 & 0.94 & 0.94\tabularnewline

\end{tabular}
    \label{tab:traj_10_picp}
    \vspace{-0.5cm}
\end{table}

\begin{table}[t]
    \centering{}\caption{\textcolor{black}{Performance of PoET on the synthetic YCB-V dataset.}}
\scalebox{0.9}{
\begin{tabular}{l|c|c|c}
Object & AUC of ADD-S & Avg. TE {[}m{]} & Avg. RE {[}°{]}\tabularnewline
\hline 
\hline 
cracker box & 79.1 & 0.038 & 34.52\tabularnewline
sugar box & 77.3 & 0.043 & 34.34\tabularnewline
mustard bottle & 79.7 & 0.039 & 37.95\tabularnewline
bleach cleanser & 78.9 & 0.038 & 26.07\tabularnewline
mug & 84.6 & 0.032 & 49.62\tabularnewline
power drill & 80.7 & 0.037 & 25.02\tabularnewline
scissors & 66.2 & 0.053 & 46.74\tabularnewline
\hline 
All & 78.1 & 0.040 & 36.06\tabularnewline
\end{tabular}
}
    \label{tab:synt_ycbv_metric}
    \vspace{-0.6cm}
\end{table}

\begin{figure*}[t]
    \centering
    \includegraphics[width=0.9\columnwidth]{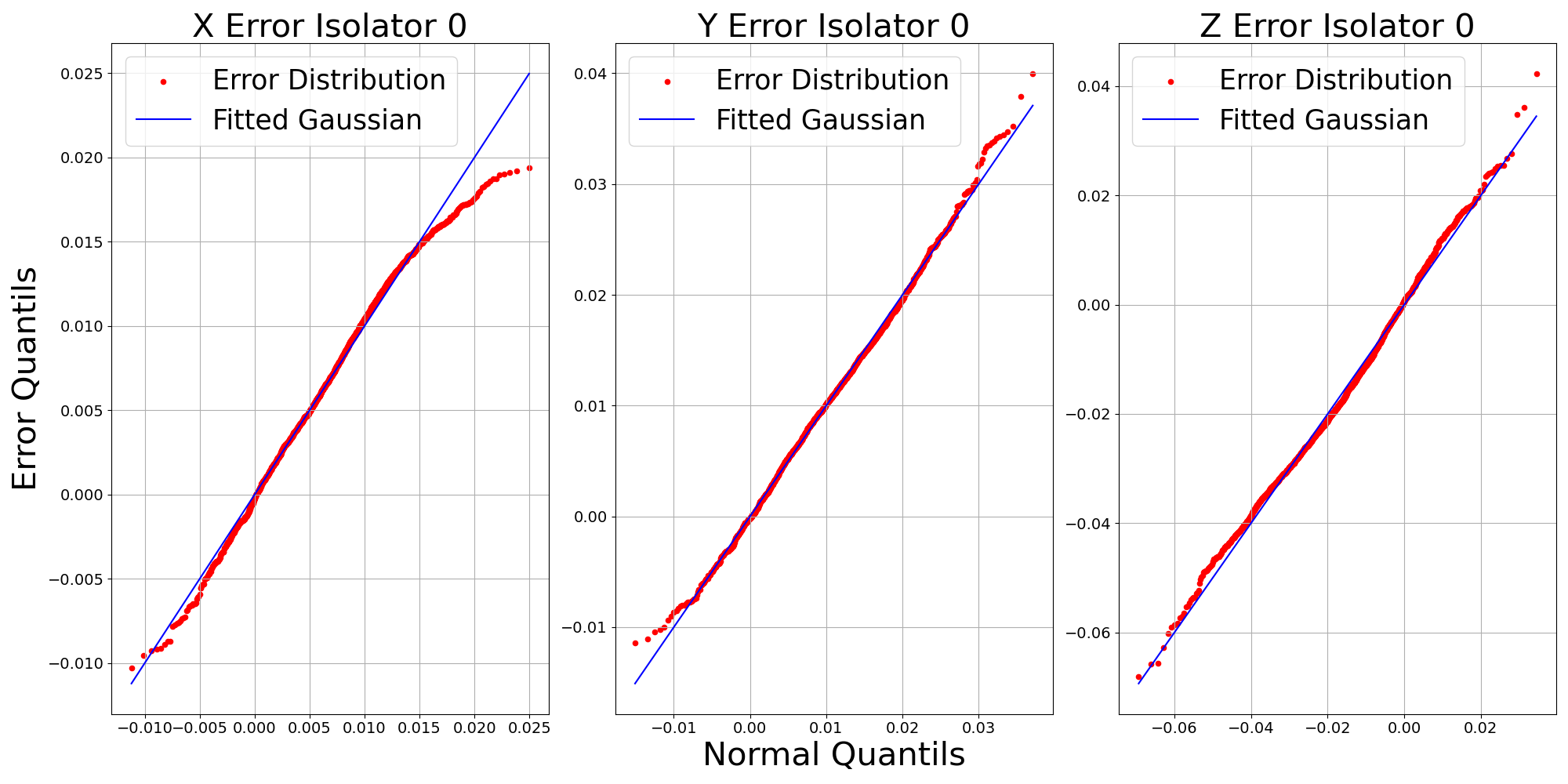}
    \includegraphics[width=0.9\columnwidth]{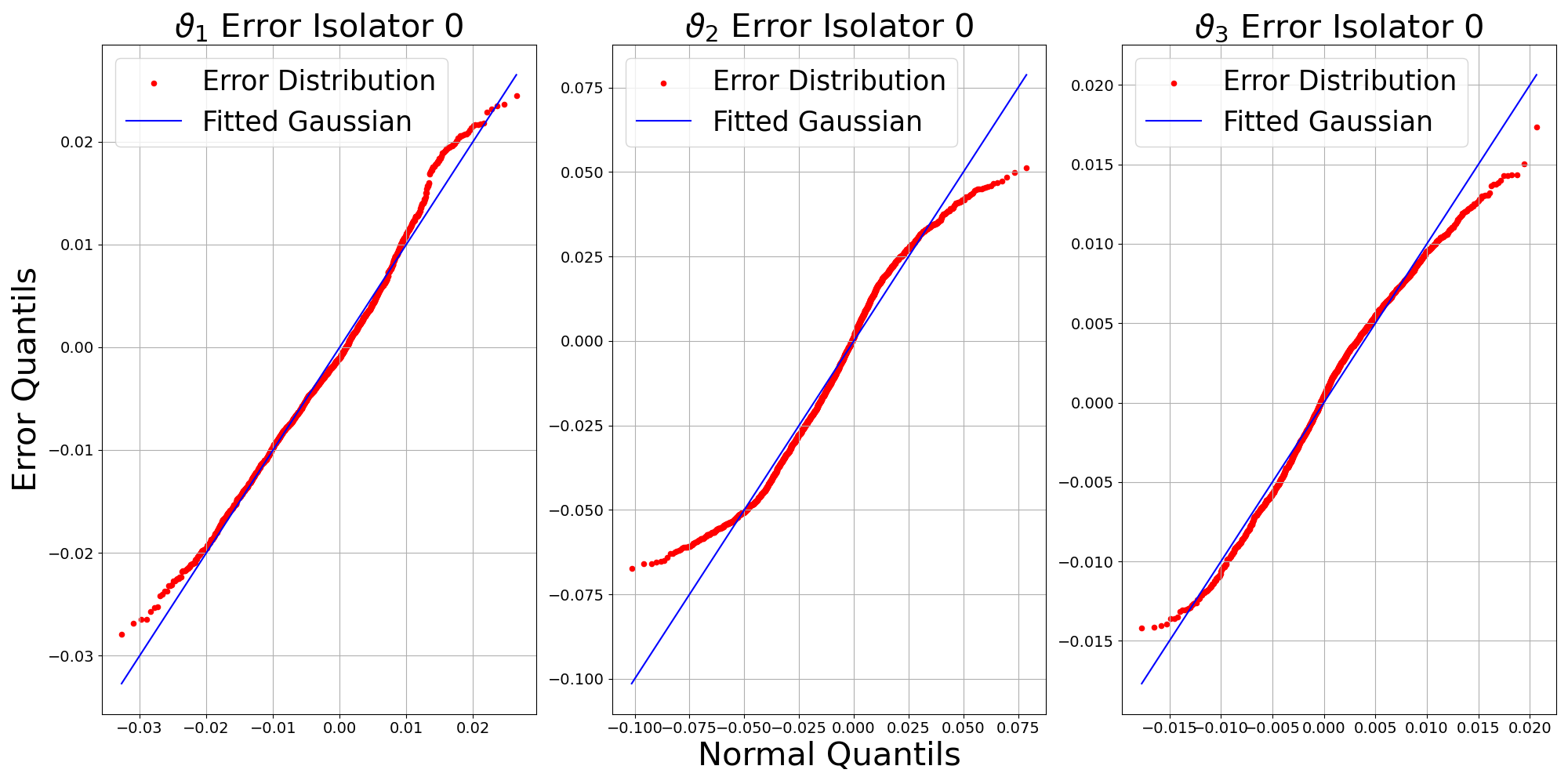}
    \vspace{-0.2cm}
    \caption{Componentwise Q-Q plot comparing the error distribution (\texttt{red}) for the translation (\texttt{left}) and the rotation (\texttt{right}) to a fitted Gaussian distribution (\texttt{blue}). The comparison is conducted for insulator 0 and the trajectory depicted in \cref{fig:example_synt_trajectory}.}
    \label{fig:qq_error_distribution}
    \vspace{-0.3cm}
\end{figure*}

Comparing the error distributions of one insulator across the whole trajectory to fitted normal distributions with the help of a Q-Q plot, cf. \cref{fig:qq_error_distribution}, underlines our assumption of a Gaussian distributed error for almost every 6D pose component. While an overall good Gaussian fit is observable for the translation error components, the $x$ error shows slight deviations at the high quantiles, indicating larger tails for the Gaussian distribution. On the other hand, slight discrepancies in the rotational components' error distribution are noticeable. Although not as pronounced as for the $x$ error, the errors of $\vartheta_1$ and $\vartheta_3$ also follow a Gaussian distribution with slight deviations towards the tails. In contrast, the error distribution of $\vartheta_2$ shows more pronounced deviations from the Gaussian distribution towards the tails and slight asymmetry of the error distribution, meaning higher positive errors are more frequent. In general, the error distributions of all 6D pose components are not exactly zero-mean, which might be attributed to the aleatoric loss function, see \cref{eq:loss_function}. As it was outlined in \cref{subsec:object_pose_estimation}, the network parameters are optimized such that the predicted uncertainty ($\hat{\sigma}_{y_{n,i}}^2$) resembles the error ($e_{n,i}^2$) of the prediction, leading to undesirable behavior of the loss function, i.e. steep gradients, when the error is tending towards zero, punishing the network for exact predictions. However, the error means are close enough to zero, so their impact on the object-relative state estimator will be minimal.

\begin{figure*}
    \centering
    \includegraphics[width=0.9\columnwidth]{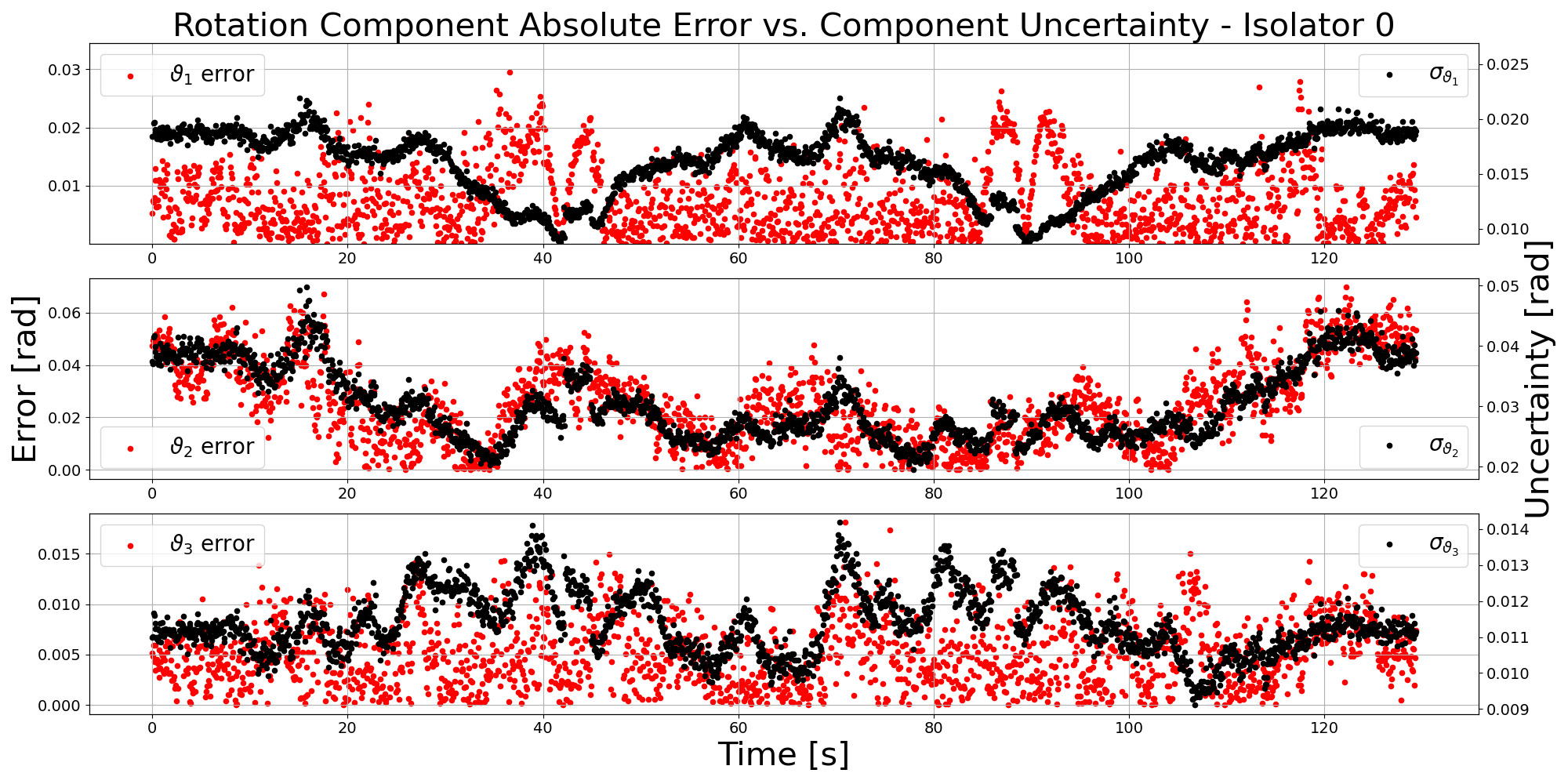}
    \includegraphics[width=0.9\columnwidth]{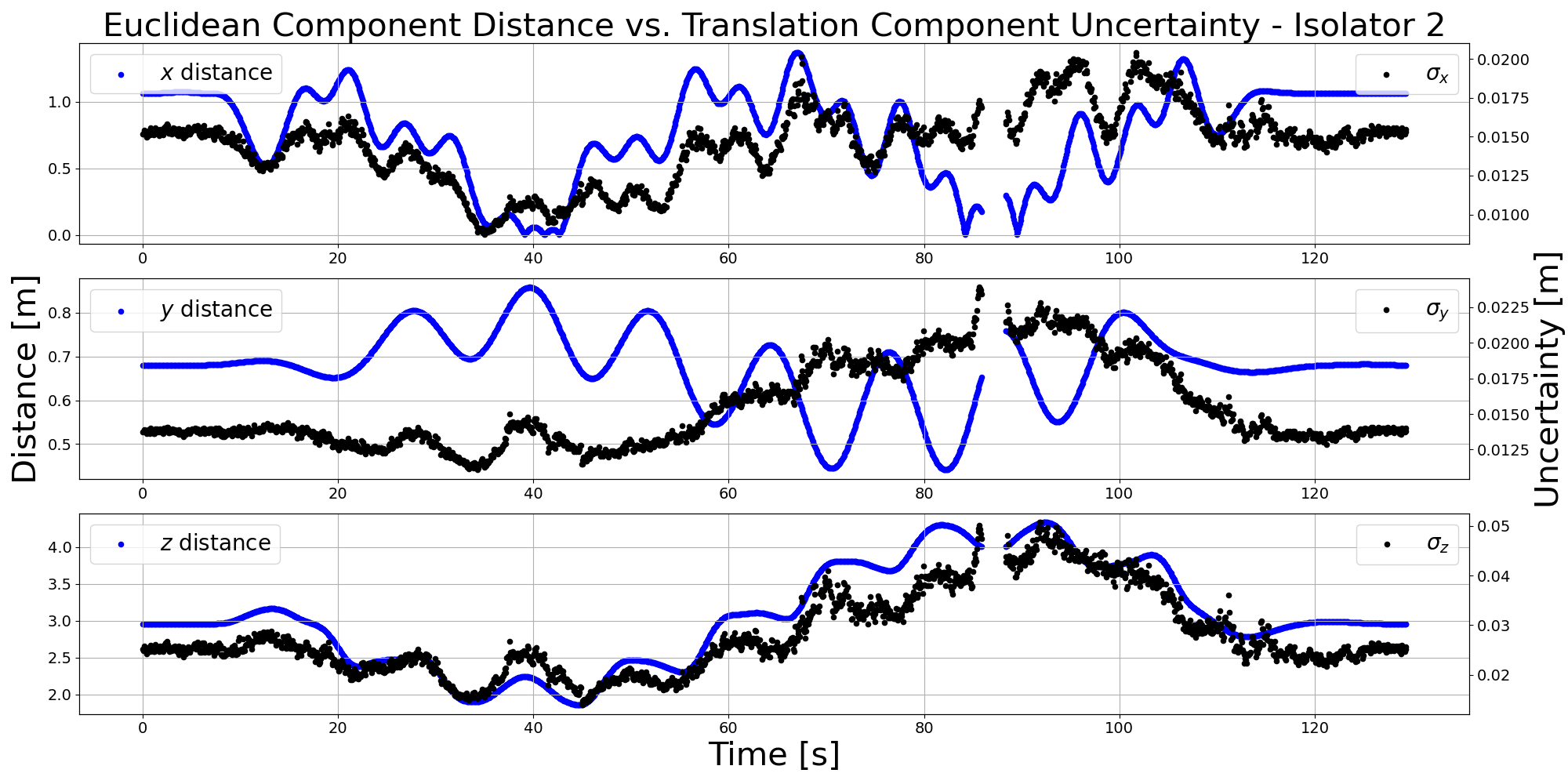}
    \vspace{-0.3cm}
    \caption{\texttt{Left}: Comparison of the absolute rotation error (\texttt{red}) to the estimated aleatoric uncertainty (\texttt{black}) across the whole trajectory for $I_0$. \texttt{Right}: Comparison of the distance (\texttt{blue}) to the estimated aleatoric uncertainty (\texttt{black}) across the whole trajectory for $I_2$. The gaps in the graphs are caused by the power pole occluding the insulator.  Note the different scales in the plots' axes (left and right sides of the plots) and across the plots.}
    \label{fig:error_distance_uncertainty}
    \vspace{-0.3cm}
\end{figure*}

In \cref{fig:error_distance_uncertainty}, the absolute rotation error is compared componentwise to the predicted aleatoric uncertainty. We focus on $I_0$ to observe the uncertainties' behavior in the case of a slightly varying distance to the camera. Except for the case of viewing the power pole from the side ($t=41s$ and $t=90s$), where the $\vartheta_1$ uncertainty tends towards zero while the error has two succinct peaks, the rotational uncertainty closely follows the error in each component. Additionally, we investigate the influence of the distance to the insulator on the uncertainty prediction for $I_2$. The bottom right plot shows the strong correlation between the predicted uncertainty and the distance to the object for the camera's $z$-axis, which is pointing out of the image plane. Even though the uncertainty in the $x$-component effectively captures local variation in the distance of the $x$-direction, particularly at close distances, the uncertainties' overall magnitude is strongly influenced by larger distances along the $z$-axis. This is especially visible in the time between $80s$ and $100s$. The distance to the object is the primary source of translation aleatoric uncertainty, which can be explained by the amount of information available regarding the object. The further away the object is, the fewer pixels capture it, providing less information to the network. Note the different scales in the plots' axes and across the plots.

\begin{figure*}
    \centering
    \includegraphics[width=0.9\columnwidth]{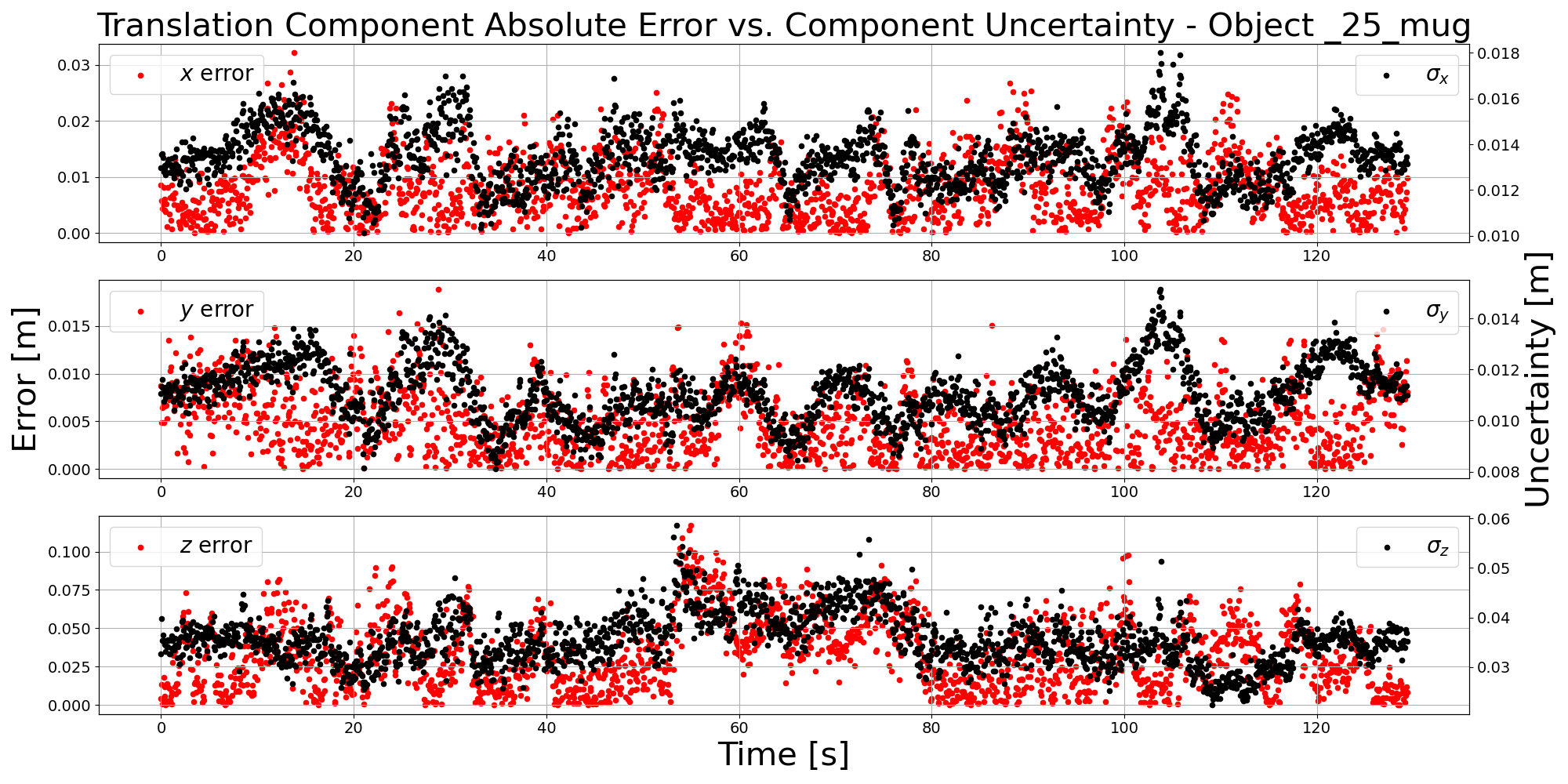}
    \includegraphics[width=0.9\columnwidth]{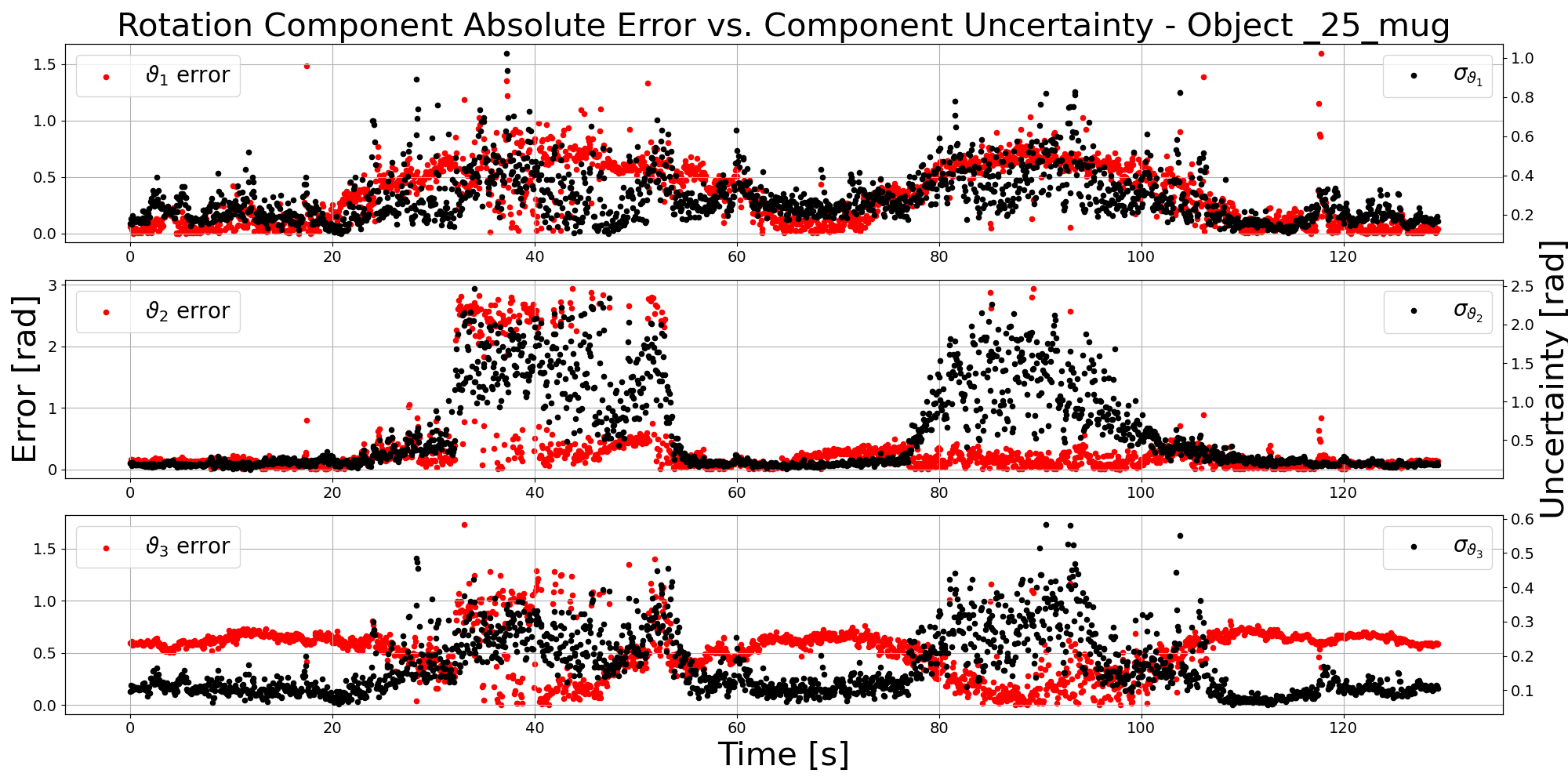}
    \vspace{-0.3cm}
    \caption{Comparison of the absolute translation error (\texttt{left}, \texttt{red}) and rotation error (\texttt{right}, \texttt{red}) to the estimated aleatoric uncertainty (\texttt{black}) across the whole trajectory for the mug. Note the different scales in the plots' axes (left and right sides of the plots) and across the plots.}
    \label{fig:ycbv_error_uncertainty}
    \vspace{-0.45cm}
\end{figure*}

For the YCB-V dataset, PoET is trained using the \texttt{Calib} scheme described above. In \cref{tab:synt_ycbv_metric}, we report the performance of PoET on the validation dataset in terms of the AUC of ADD-S \cite{xiang2018posecnn}, average translation and rotation error. As for the power pole, the predicted uncertainty is analyzed by placing each object at the center of a synthetic trajectory with a fixed distance of 1.0m. The PICP scores are reported in \cref{tab:synt_ycbv_picp}. Except for the $z$-component of some objects, the predicted aleatoric uncertainty captures the errors of the translation and rotation components. The only outliers are the rotational components of the mug and the scissors, which are almost symmetrical objects. In \cref{fig:ycbv_error_uncertainty}, the absolute translation and rotation errors are compared to the predicted aleatoric uncertainty of the mug. While the uncertainty nicely encapsulates the translation components, the rotation components exhibit heightened error and uncertainty values for ambiguous scenarios, i.e., where the handle is not clearly visible. Given this observation, performing aleatoric uncertainty-based outlier rejection (\texttt{AOR}) is possible by determining suitable thresholds.

\begin{table}[t]
    \caption{\textcolor{black}{Comparison of the PICP score for each object for $\alpha=0.95$}}
\centering{}%
\begin{tabular}{l|c|c|c||c|c|c}
Object & x & y & z & $\vartheta_1$ & $\vartheta_2$ & $\vartheta_3$\tabularnewline
\hline 
\hline 
cracker box & 0.987 & 0.995 & 0.866 & 0.979 & 0.905 & 0.954\tabularnewline
sugar box & 0.990 & 0.998 & 0.869 & 0.969 & 0.885 & 0.992\tabularnewline
mustard bottle & 0.994 & 0.999 & 0.842 & 0.975 & 0.961 & 0.993\tabularnewline
bleach cleanser & 0.996 & 0.998 & 0.997 & 0.924 & 0.922 & 0.988\tabularnewline
mug & 0.982 & 1.000 & 0.862 & 0.703 & 0.953 & 0.228\tabularnewline
power drill & 0.996 & 0.993 & 0.962 & 0.970 & 0.941 & 0.992\tabularnewline
scissors & 0.988 & 0.998 & 0.901 & 0.865 & 0.831 & 0.313\tabularnewline
\end{tabular}
    \label{tab:synt_ycbv_picp}
    \vspace{-0.6cm}
\end{table}

The results indicate that our proposed network can capture the metric aleatoric uncertainty of the predicted 6D object pose. As modeled in \cref{subsec:object_pose_estimation}, the translation and the rotation uncertainty can be predicted componentwise and capture the characteristics of the error, which are Gaussian distributed.
\looseness=-1
\vspace{-0.4cm}
\subsection{Object-relative State Estimation}
\label{subsec:eval_state_estimation}

The in-depth analysis above shows that the predicted aleatoric uncertainty adequately represents the potential error of the predicted, relative 6D object pose. This section illustrates that these aleatoric uncertainties can be used as a dynamic measurement uncertainty in object-relative state estimation. This eliminates the need for expert knowledge and manual tuning and also enables dynamic anchor object switching benefiting the object-relative state estimation task.

Following \cref{subsec:dataset}, we generated five trajectories with varying distances, heights, and durations to evaluate the influence of the measurement uncertainty on the performance of the object-relative state estimator. We compare three different approaches for determining and utilizing the measurement uncertainty $\bm{\Sigma}_t$ and $\bm{\Sigma}_\vartheta$: First, following the idea of \cite{zorina2025temporally} to empirically determine the measurement uncertainty, the average error on the validation dataset is calculated and the standard deviation for each component is set to the corresponding error, i.e., $\sigma_{x,y,z} = 0.03$m and $\sigma_{\vartheta_{1,2,3}} = 0.035$rad. This measurement noise covariance is kept fixed throughout the estimation (\texttt{Fixed}). Second, the predicted aleatoric uncertainties are used to construct the measurement uncertainty as described in \cref{subsec:state_estimation}. Third, dynamic anchor object switching is enabled based on the dynamic measurement noise derived from the aleatoric uncertainty. The predicted 6D object poses from the \texttt{Calib} network are used as measurements for each approach. In \cref{tab:synthetic_data}, the state estimator's performance is evaluated based on the root mean square error (RMSE) for position and orientation, as well as the maximum position error (PE). Compared to empirically determined uncertainty (left columns), the aleatoric measurement uncertainty (middle columns) achieves a comparable performance for all three metrics but without the need for prior empirical estimation. Including dynamic switching of the anchor object, enabled only through the aleatoric uncertainty prediction per measurement, drastically improves the performance for object-relative state estimation (right columns). This highlights the benefits of aleatoric uncertainty for DL-based object-relative state estimation.
\looseness=-1

To verify its real world applicability, we evaluate our approach on the real-world UAV data from \cite{jantos2024aivio}\footnote{Online available at \href{https://www.aau.at/en/smart-systems-technologies/control-of-networked-systems/datasets/aivio/}{AIVIO dataset homepage}.}. It mimics an inspection flight around a power pole by flying on a $\pm50^\circ$ arc at a fixed distance of 3.3m. The data includes real IMU data at 200 Hz, RGB images at 15 Hz, and ground truth data at 60 Hz from a motion capture system. The provided noise values are not changed. The results are reported in \cref{tab:aivio_data}. As in the case of the synthetic data, the state estimator achieves a similar performance for the fixed and aleatoric uncertainty-based measurement uncertainty, while the dynamic anchor object switching once again leads to improvements. The predicted aleatoric uncertainty of the 6D object pose predictor, solely trained on synthetic data, also benefits the state estimation in real-world scenarios.
\looseness=-1

For the YCB-V objects, we simulate five synthetic trajectories with different object constellations. Once again, the fixed measurement noise is determined by the average error of PoET reported in \cref{tab:synt_ycbv_metric}, i.e., $\sigma_{x,y,z} = 0.04$m and $\sigma_{\vartheta_{1,2,3}} = 0.628$rad. For comparison, the aleatoric uncertainty is combined with \texttt{AOR}. We report the RMSE for position and orientation in \cref{tab:synt_ycbv_trajectory}. Aleatoric uncertainty and \texttt{AOR} improve the state estimation performance.
\looseness=-1

\begin{table*}
    \centering\caption{Results for object-relative state estimation using synthetic trajectories. We compare fixed measurement uncertainty with our proposed aleatoric uncertainty and dynamic anchor object switching. Bold values highlight best results.}
\scalebox{0.8}{
\begin{tabular}{c|c|c|c|c|c|c|c|c|c}
\multirow{2}{*}{Trajectory} & \multicolumn{3}{c|}{Fixed} & \multicolumn{3}{c|}{Aleatoric Uncertainty (AU)} & \multicolumn{3}{c}{AU + Dynamic Anchor Object Switching}\tabularnewline
\cline{2-10} \cline{3-10} \cline{4-10} \cline{5-10} \cline{6-10} \cline{7-10} \cline{8-10} \cline{9-10} \cline{10-10} 
 & RMSE{[}m{]} & RMSE{[}°{]} & Max PE {[}m{]} & RMSE{[}m{]} & RMSE{[}°{]} & Max PE {[}m{]} & RMSE{[}m{]} & RMSE{[}°{]} & Max PE {[}m{]}\tabularnewline
\hline 
\hline 
1 & 0.142 & 2.44 & 0.305 & 0.143 & 2.46 & \textbf{0.302} & \textbf{0.128} & \textbf{2.26} & 0.304\tabularnewline
2 & 0.136 & 2.33 & 0.297 & 0.137 & 2.36 & 0.296 & \textbf{0.130} & \textbf{2.27} & \textbf{0.261}\tabularnewline
3 & \textbf{0.118} & \textbf{2.05} & 0.292 & \textbf{0.118} & \textbf{2.05} & 0.292 & 0.119 & 2.13 & \textbf{0.284}\tabularnewline
4 & 0.160 & 2.93 & 0.347 & 0.161 & 2.94 & 0.347 & \textbf{0.121} & \textbf{2.12} & \textbf{0.257}\tabularnewline
5 & 0.143 & 2.31 & 0.342 & 0.147 & 2.37 & 0.375 & \textbf{0.130} & \textbf{2.10} & \textbf{0.295}\tabularnewline
\hline 
Mean & 0.140 & 2.41 & 0.317 & 0.141 & 2.44 & 0.322 & \textbf{0.126} & \textbf{2.18} & \textbf{0.280}\tabularnewline
\hline 
\end{tabular}
}
    \label{tab:synthetic_data}
    \vspace{-0.3cm}
\end{table*}
\begin{table*}
    \centering
\caption{Results for object-relative state estimation using real-world data \cite{jantos2024aivio}. We compare fixed measurement uncertainty with our proposed aleatoric uncertainty and dynamic anchor object switching. Bold values highlight best results.}
\vspace{-0.2cm}
\scalebox{0.8}{
\begin{tabular}{c|c|c|c|c|c|c|c|c|c}
\multirow{2}{*}{Flight} & \multicolumn{3}{c|}{Fixed} & \multicolumn{3}{c|}{Aleatoric Uncertainty (AU)} & \multicolumn{3}{c}{AU + Dynamic Anchor Object Switching}\tabularnewline
\cline{2-10} \cline{3-10} \cline{4-10} \cline{5-10} \cline{6-10} \cline{7-10} \cline{8-10} \cline{9-10} \cline{10-10} 
 & RMSE{[}m{]} & RMSE{[}°{]} & Max PE {[}m{]} & RMSE{[}m{]} & RMSE{[}°{]} & Max PE {[}m{]} & RMSE{[}m{]} & RMSE{[}°{]} & Max PE {[}m{]}\tabularnewline
\hline 
\hline 
1 & 0.176 & 2.47 & 0.350 & 0.175 & 2.40 & 0.359 & \textbf{0.151}& \textbf{2.09} & \textbf{0.293}\tabularnewline
2 & 0.177 & 2.00 & 0.372 & 0.193 & 2.02 & 0.381 & \textbf{0.140} & \textbf{1.72} & \textbf{0.315}\tabularnewline
3 & 0.180 & 2.52 & 0.356 & 0.187 & 2.61 & 0.373 & \textbf{0.128} & \textbf{1.93} & \textbf{0.257}\tabularnewline
4 & 0.163 & 2.00 & 0.314 & 0.170 & 2.05 & 0.335 & \textbf{0.139} & \textbf{1.80} & \textbf{0.277}\tabularnewline
5 & 0.165 & \textbf{2.09} & 0.312 & 0.169 & 2.10 & 0.313 & \textbf{0.148} & 2.18 & \textbf{0.291}\tabularnewline
6 & 0.178 & 2.27 & 0.375 & 0.187 & 2.19 & 0.394 & \textbf{0.138} & \textbf{1.83} & \textbf{0.334}\tabularnewline
7 & \textbf{0.167} & \textbf{2.10} & \textbf{0.350} & 0.174 & \textbf{2.10} & 0.352 & 0.180 & 2.52 & 0.355\tabularnewline
\hline 
Mean & 0.172 & 2.21 & 0.347 & 0.179 & 2.21 & 0.358 & \textbf{0.146} & \textbf{2.01} & \textbf{0.303}\tabularnewline
\hline 
\end{tabular}
}
    \label{tab:aivio_data}
    \vspace{-0.4cm}
\end{table*}
\begin{table}
    \centering
\caption{\textcolor{black}{Results for object-relative state estimation using synthetic YCB-V data. Bold values highlight best results.}}
\scalebox{0.8}{
\begin{tabular}{c|c|c|c|c}
\multirow{2}{*}{} & \multicolumn{2}{c|}{Fixed} & \multicolumn{2}{c}{AU + AOR}\tabularnewline
\cline{2-5} \cline{3-5} \cline{4-5} \cline{5-5} 
 & RMSE{[}m{]} & RMSE{[}°{]} & RMSE{[}m{]} & RMSE{[}°{]}\tabularnewline
\hline 
\hline 
1 & 0.593 & 34.31 & \textbf{0.209} & \textbf{7.18}\tabularnewline
2 & \textbf{0.182} & 20.06 & 0.201 & \textbf{8.23}\tabularnewline
3 & 0.567 & 18.38 & \textbf{0.513} & \textbf{15.06}\tabularnewline
4 & 0.516 & 28.94 & \textbf{0.383} & \textbf{17.51}\tabularnewline
5 & 0.265 & 25.82 & \textbf{0.248} & \textbf{10.66}\tabularnewline
\hline 
Mean & 0.425 & 25.50 & \textbf{0.311} & \textbf{11.73}\tabularnewline
\hline 
\end{tabular}
}
    \label{tab:synt_ycbv_trajectory}
    \vspace{-0.6cm}
\end{table}
\vspace{-0.4cm}
\section{Conclusion} \label{sec:concluison}
In this letter, we presented an approach to extend a (pre-trained) DL-based 6D object pose predictor for aleatoric uncertainty prediction of the full 6D pose, modeled as a multivariate Gaussian. The predicted metric aleatoric uncertainty captures the error characteristics of the predicted 6D pose and is thus a well suited uncertainty measure for down-stream robotics tasks and can be used for outlier rejection (\texttt{AOR}). Concretely, we showed that utilizing the predicted aleatoric uncertainty as dynamic measurement covariance leads to improvements in the object-relative state estimation task. Besides that, it removes the need for expert knowledge or tedious empirical experiments to determine a suitable measurement covariance. The predicted uncertainty per measurement also allowed a stochastic decision process to dynamically determine the object to which the navigation frame is anchored to. This is not possible with fix uncertainty approaches and led to significant state estimation improvements in both synthetic and real experiments.
We also highlighted that the 6D object pose predictor including the aleatoric uncertainty heads can be trained solely on synthetic data with its performance translating to real-world data. Moreover, it was shown that it is sufficient to freeze the pre-trained object detector and calibrate the aleatoric uncertainty heads to achieve performance improvements for the state estimation task. 
\vspace{-0.4cm}
\bibliographystyle{IEEEtran.bst}
\bibliography{IEEEabrv, root.bib}

\begin{thebibliography}{10}
\providecommand{\url}[1]{#1}
\csname url@rmstyle\endcsname
\providecommand{\newblock}{\relax}
\providecommand{\bibinfo}[2]{#2}
\providecommand\BIBentrySTDinterwordspacing{\spaceskip=0pt\relax}
\providecommand\BIBentryALTinterwordstretchfactor{4}
\providecommand\BIBentryALTinterwordspacing{\spaceskip=\fontdimen2\font plus
\BIBentryALTinterwordstretchfactor\fontdimen3\font minus \fontdimen4\font\relax}
\providecommand\BIBforeignlanguage[2]{{%
\expandafter\ifx\csname l@#1\endcsname\relax
\typeout{** WARNING: IEEEtran.bst: No hyphenation pattern has been}%
\typeout{** loaded for the language `#1'. Using the pattern for}%
\typeout{** the default language instead.}%
\else
\language=\csname l@#1\endcsname
\fi
#2}}

\bibitem{jantos2024aivio}
T.~Jantos, M.~Scheiber, C.~Brommer, E.~Allak, S.~Weiss, and J.~Steinbrener, ``Aivio: Closed-loop, object-relative navigation of uavs with ai-aided visual inertial odometry,'' \emph{IEEE Robotics and Automation Letters}, vol.~9, no.~12, pp. 10\,764--10\,771, 2024.

\bibitem{jantos2023ai}
T.~Jantos, C.~Brommer, E.~Allak, S.~Weiss, and J.~Steinbrener, ``Ai-based multi-object relative state estimation with self-calibration capabilities,'' in \emph{2023 IEEE International Conference on Robotics and Automation (ICRA)}, 2023, pp. 2789--2795.

\bibitem{gawlikowski2023aisurvey}
J.~Gawlikowski, C.~R.~N. Tassi, M.~Ali, J.~Lee, M.~Humt, J.~Feng, A.~Kruspe, R.~Triebel, P.~Jung, R.~Roscher, \emph{et~al.}, ``A survey of uncertainty in deep neural networks,'' \emph{Artificial Intelligence Review}, vol.~56, no. Suppl 1, pp. 1513--1589, 2023.

\bibitem{kendall2017uncertainties}
A.~Kendall and Y.~Gal, ``What uncertainties do we need in bayesian deep learning for computer vision?'' \emph{Advances in neural information processing systems}, vol.~30, 2017.

\bibitem{sato2023probabilistic}
H.~Sato, T.~Ikeda, and K.~Nishiwaki, ``A probabilistic rotation representation for symmetric shapes with an efficiently computable bingham loss function,'' in \emph{2023 IEEE International Conference on Robotics and Automation (ICRA)}.\hskip 1em plus 0.5em minus 0.4em\relax IEEE, 2023, pp. 6923--6929.

\bibitem{haugaard2023spyropose}
R.~L. Haugaard, F.~Hagelskj{\ae}r, and T.~M. Iversen, ``Spyropose: Se (3) pyramids for object pose distribution estimation,'' in \emph{Proceedings of the IEEE/CVF International Conference on Computer Vision}, 2023.

\bibitem{wursthorn2024uncertainty}
K.~Wursthorn, M.~Hillemann, and M.~Ulrich, ``Uncertainty quantification with deep ensembles for 6d object pose estimation,'' \emph{ISPRS Annals of the Photogrammetry, Remote Sensing and Spatial Information Sciences}, 2024.

\bibitem{merrill2022symmetry}
N.~Merrill, Y.~Guo, X.~Zuo, X.~Huang, S.~Leutenegger, X.~Peng, L.~Ren, and G.~Huang, ``Symmetry and uncertainty-aware object slam for 6dof object pose estimation,'' in \emph{Proceedings of the IEEE/CVF Conference on Computer Vision and Pattern Recognition}, 2022, pp. 14\,901--14\,910.

\bibitem{zorina2025temporally}
K.~Zorina, V.~Priban, M.~Fourmy, J.~Sivic, and V.~Petrik, ``Temporally consistent object 6d pose estimation for robot control,'' \emph{IEEE Robotics and Automation Letters}, vol.~10, no.~1, pp. 56--63, 2025.

\bibitem{yang2020d3vo}
N.~Yang, L.~v. Stumberg, R.~Wang, and D.~Cremers, ``D3vo: Deep depth, deep pose and deep uncertainty for monocular visual odometry,'' in \emph{Proceedings of the IEEE/CVF conference on computer vision and pattern recognition}, 2020, pp. 1281--1292.

\bibitem{han2024nvins}
J.~Han, L.~L. Beyer, G.~V. Cavalheiro, and S.~Karaman, ``Nvins: Robust visual inertial navigation fused with nerf-augmented camera pose regressor and uncertainty quantification,'' in \emph{IEEE/RSJ International Conference on Intelligent Robots and Systems (IROS)}, 2024.

\bibitem{peretroukhin2019probabilistic}
V.~Peretroukhin, B.~Wagstaff, M.~Giamou, and J.~Kelly, ``Probabilistic regression of rotations using quaternion averaging and a deep multi-headed network,'' \emph{arXiv preprint arXiv:1904.03182}, 2019.

\bibitem{russell2021multivariate}
R.~L. Russell and C.~Reale, ``Multivariate uncertainty in deep learning,'' \emph{IEEE Transactions on Neural Networks and Learning Systems}, vol.~33, no.~12, pp. 7937--7943, 2021.

\bibitem{moreau2022coordinet}
A.~Moreau, N.~Piasco, D.~Tsishkou, B.~Stanciulescu, and A.~de~La~Fortelle, ``Coordinet: uncertainty-aware pose regressor for reliable vehicle localization,'' in \emph{Proceedings of the IEEE/CVF Winter Conference on Applications of Computer Vision}, 2022.

\bibitem{jantos2023poet}
T.~Jantos, M.~A. Hamdad, W.~Granig, S.~Weiss, and J.~Steinbrener, ``{PoET: Pose Estimation Transformer for Single-View, Multi-Object 6D Pose Estimation},'' in \emph{Proceedings of the 6th Conference on Robot Learning}.\hskip 1em plus 0.5em minus 0.4em\relax PMLR, 2023.

\bibitem{zhou2019continuity}
Y.~Zhou, C.~Barnes, J.~Lu, J.~Yang, and H.~Li, ``On the continuity of rotation representations in neural networks,'' in \emph{Proceedings of the IEEE/CVF Conference on Computer Vision and Pattern Recognition}, 2019, pp. 5745--5753.

\bibitem{brommer2020mars}
C.~Brommer, R.~Jung, J.~Steinbrener, and S.~Weiss, ``Mars: A modular and robust sensor-fusion framework,'' \emph{IEEE Robotics and Automation Letters}, vol.~6, no.~2, pp. 359--366, 2020.

\bibitem{sola2017quaternion}
J.~Sola, ``Quaternion kinematics for the error-state kalman filter,'' \emph{arXiv preprint arXiv:1711.02508}, 2017.

\bibitem{xiang2018posecnn}
Y.~Xiang, T.~Schmidt, V.~Narayanan, and D.~Fox, ``Posecnn: A convolutional neural network for 6d object pose estimation in cluttered scenes,'' in \emph{Robotics: Science and Systems (RSS)}, 2018.

\bibitem{wang2021scaledyolo}
C.-Y. Wang, A.~Bochkovskiy, and H.-Y.~M. Liao, ``{Scaled-YOLOv4}: Scaling cross stage partial network,'' in \emph{Proceedings of the IEEE/CVF Conference on Computer Vision and Pattern Recognition (CVPR)}, June 2021, pp. 13\,029--13\,038.

\end{thebibliography}
\end{document}